\documentclass[letterpaper]{article} 
\usepackage{aaai25}  
\usepackage{times}  
\usepackage{helvet}  
\usepackage{courier}  
\usepackage[hyphens]{url}  
\usepackage{graphicx} 
\usepackage{natbib}  
\usepackage{caption} 
\usepackage{algorithm}
\usepackage{algorithmic}
\usepackage{subfigure}
\usepackage{pdfpages}
\usepackage{multicol}
\usepackage{newfloat}
\usepackage{listings}
\DeclareCaptionStyle{ruled}{labelfont=normalfont,labelsep=colon,strut=off} 
\floatstyle{ruled}
\newfloat{listing}{tb}{lst}{}
\floatname{listing}{Listing}
\usepackage{amsmath}
\usepackage{amssymb}
\usepackage{mathtools}
\usepackage{amsthm}
\usepackage{multicol}
\usepackage{multirow}
\usepackage{microtype}
\usepackage{graphicx}
\usepackage{booktabs} 
\usepackage{color}
\newtheorem{theorem}{Theorem}

\usepackage{verbatim}

\title{One Node One Model: Featuring the Missing-Half for Graph Clustering}
\author{Xuanting Xie\textsuperscript{\rm 1}, Bingheng Li\textsuperscript{\rm 2}, Erlin Pan\textsuperscript{\rm 3}, Zhaochen Guo\textsuperscript{\rm 1}, Zhao Kang\textsuperscript{\rm 1}\thanks{Corresponding author.}, Wenyu Chen\textsuperscript{\rm 1}
}

\affiliations{
    \textsuperscript{\rm 1}University of Electronic Science and Technology of China, Chengdu, Sichuan, China \\
    \textsuperscript{\rm 2}Michigan State University, East Lansing, US\\
    \textsuperscript{\rm 3}Alibaba Group \\
    x624361380@outlook.com, libinghe@msu.edu, panerlin.pel@alibaba-inc.com, \{zkang, cwy\}@uestc.edu.cn
}

\usepackage{bibentry}

\begin{document}

\maketitle

\begin{abstract}
Most existing graph clustering methods primarily focus on exploiting topological structure, often neglecting the ``missing-half" node feature information, especially how these features can enhance clustering performance. This issue is further compounded by the challenges associated with high-dimensional features. Feature selection in graph clustering is particularly difficult because it requires simultaneously discovering clusters and identifying the relevant features for these clusters. To address this gap, we introduce a novel paradigm called ``one node one model", which builds an exclusive model for each node and defines the node label as a combination of predictions for node groups. Specifically, the proposed ``Feature Personalized Graph Clustering (FPGC)" method identifies cluster-relevant features for each node using a squeeze-and-excitation block, integrating these features into each model to form the final representations. Additionally, the concept of feature cross is developed as a data augmentation technique to learn low-order feature interactions. Extensive experimental results demonstrate that FPGC outperforms state-of-the-art clustering methods. Moreover, the plug-and-play nature of our method provides a versatile solution to enhance GNN-based models from a feature perspective.
\end{abstract}

\section{Introduction}
As a fundamental task in graph data mining, attributed graph clustering aims to partition nodes into different clusters without labeled data. It is receiving increasing research attention due to the success of Graph Neural Networks (GNNs) \cite{GCN,Pcconv,qian2024upper}. Typically, GNNs use many graph convolutional layers to learn node representations by aggregating neighbor node features into the node. In recent years, many graph clustering techniques have achieved promising performance \cite{DGCN, SCGC, CDC, shen2024beyond}.

Most of these methods apply a self-supervised learning framework to fully explore the graph structure \cite{DyFSS, DMGNC}, often neglecting the ``missing-half" of node feature information, i.e., how node features can enhance clustering quality. Consequently, their performance heavily depends on the quality of the input graph. If the original graph is of poor quality, such as having many nodes from different classes connected together and missing important links, the representations learned through multilayer aggregation become nondiscrimination. This issue is known as representation collapse \cite{chen2024deep}.

\begin{figure*}[t]
    \centering
    \subfigure[Feature distribution of different clusters on Cora.]{
        \includegraphics[width=0.4\textwidth]{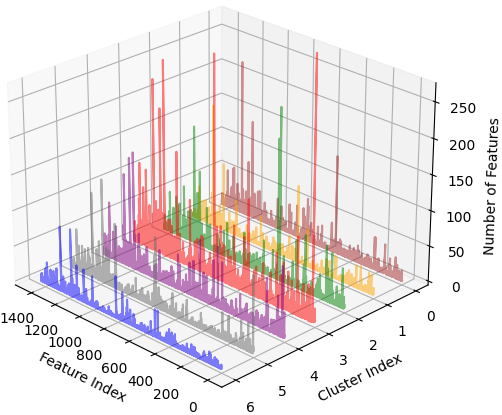}
    }
        \hspace{0.05\textwidth}
    \subfigure[DTW distance matrix with only cluster-relevant features on Cora.]{
        \includegraphics[width=0.4\textwidth]{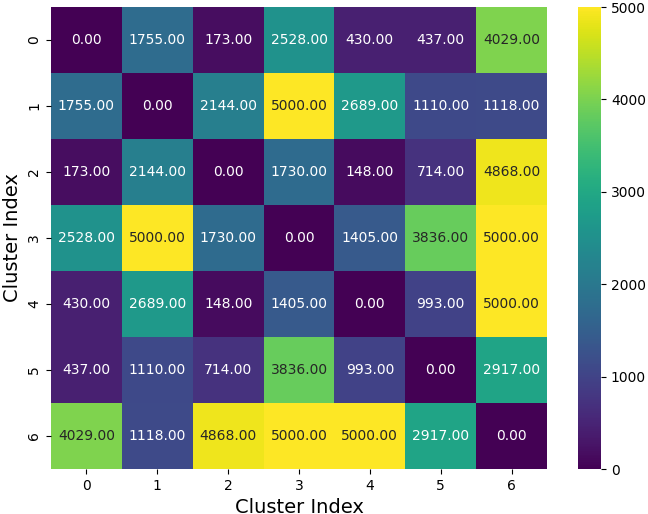}
    }
    \caption{Visualization of results on Cora. (a) is the feature distribution of different clusters, which shows that clusters are characterized by different features. (b) is the DTW distance matrix based on cluster-relevant features, which verify their distinctiveness. We can draw the conclusion that the cluster-relevant features contain valuable information about clusters.}
    \label{poso0}
\end{figure*}

For the attributed graph, the node feature and the topological structure should play an equal role in the unsupervised graph embedding and downstream clustering. 
We argue that a node's personalized features should be crucial for identifying its cluster label. For example, the Cora dataset contains 2,708 papers, each described by a 1,433-dimensional binary feature indicating the presence of keywords. We visualize the feature distributions for each cluster in Fig. \ref{poso0}(a). It can be seen that each cluster exhibits its own dominant features, while other features may be irrelevant or less important. The cluster-relevant features are the collection of all personalized features. To distinguish between clusters, we consider features that are 20 times greater than the mean value in each cluster and apply the dynamic time warping (DTW) technique \cite{muller2007dynamic} to each cluster. DTW is a similarity measure for sequences of different lengths, and we set the maximum distance to 5,000 for better visualization. From Fig. \ref{poso0}(b), we can see that there are significant differences in the cluster-relevant features. Thus, these personalized features are representative of different clusters.

Furthermore, most GNN-based methods capture only higher-order feature interactions and tend to overlook low-order interactions \cite{kim2024explicit}. The graph filtering process transforms input features into higher-level hidden representations. Recent studies on GNNs focus mainly on designing node aggregation mechanisms that emphasize various connection properties, such as local similarity \cite{GCN}, structural similarity \cite{donnat2018learning}, and multi-hop connectivity \cite{Global-hete}. However, the cross feature information is always ignored, despite its importance for model expressiveness \cite{GCNCross}. For example, by crossing a paper's keywords in the Cora dataset, such as \{$title=autonomous$ \& $approach=Q-learning$ \& $experiment=simulation$\}, the model can achieve a better paper representation and produce more accurate clustering results (e.g., this paper is categorized under Reinforcement Learning). Until recently, most existing clustering models have struggled to capture such low-order feature interactions.


To address these shortcomings, we introduce ``Feature Personalized Graph Clustering (FPGC)". We propose a novel paradigm ``one node one model" to learn a personalized model for each node, with a squeeze-and-excitation block selecting cluster-relevant features. A new data augmentation technique based on feature cross is developed to effectively capture low-order feature interactions. In summary, our contributions are as follows.


\begin{itemize}
    \item{Orthogonal to existing works, we tackle graph clustering from a feature perspective. By employing a squeeze-and-excitation block, we effectively select cluster-relevant features and propose the ``one node one model" paradigm. }
    \item{We develop a novel data augmentation technique based on feature cross to capture low-order feature interactions.}
    \item{Extensive experiments on benchmark datasets demonstrate the superiority of our method. Notably, our approach can function as a plug-and-play tool for existing GNN-based models, rather than serving solely as a stand-alone method. }
\end{itemize}

\begin{figure*}[t]
    \centering
    \includegraphics[width=.75\linewidth]{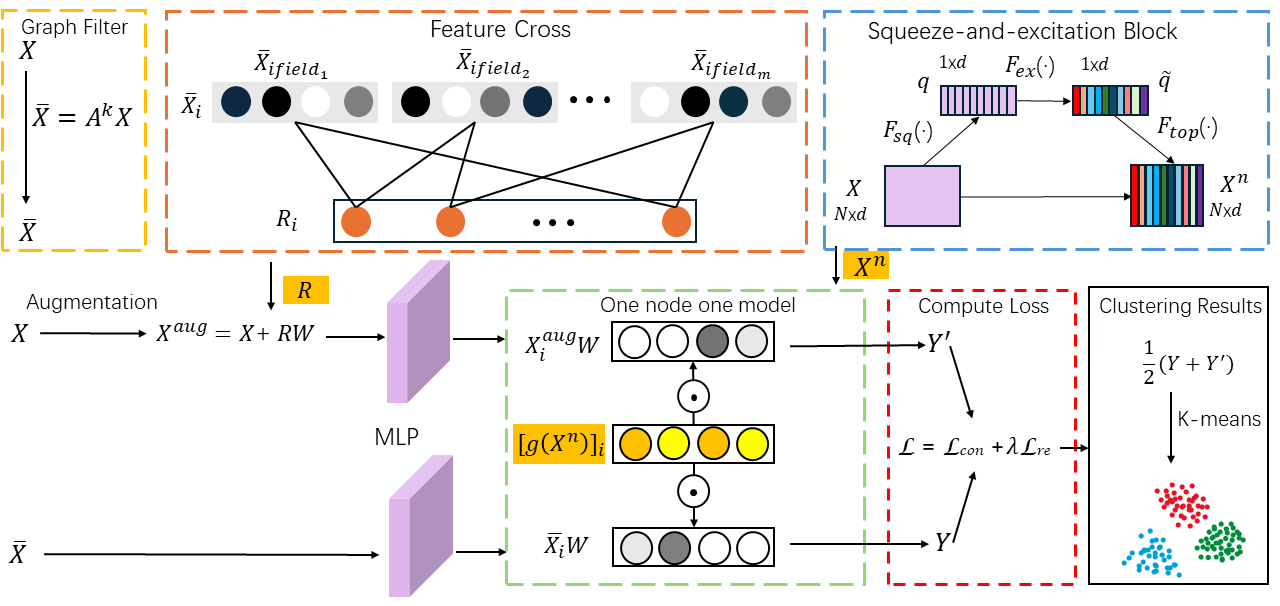}
    \caption{The pipeline of our FPGC. We preprocess the node features by stacked graph filters. Besides, we also input original features into the squeeze-and-excitation block to select the top $n$ significant features, based on which we learn a model for each node. Then, the contrastive framework encodes the smoothed node features and augmented features to achieve discriminative node representations.}
    \label{all2}
\end{figure*}

\section{Related work}
Graph clustering can be roughly divided into three categories \cite{xie2024provable,li2024simplified}. (1) Shallow methods. FGC \cite{FGC} incorporates higher-order information into graph structure learning for clustering. MCGC \cite{MCGC} uses simplified graph filtering rather than GNNs to obtain a smooth representation. (2) Contrastive methods. SCGC \cite{SCGC} simplifies network architecture and data augmentation. CCGC \cite{CCGC} leverages high-confidence clustering information to improve sample quality. CONVERT \cite{convert} uses a perturb-recover network and semantic loss for reliable data augmentation. (3) Autoencoder methods. SDCN \cite{SDCN} transfers learned representations from auto-encoders to GNN layers, employing data structure information. 
 AGE \cite{AGE} uses Laplacian smoothing filters and adaptive learning to improve node embeddings. DGCN \cite{DGCN} uses an adaptive filter to capture important frequency information and reconstructs heterophilic and homophilic graphs to handle real-world graphs with different levels of homophily. DMGNC \cite{DMGNC} uses a masked autoencoder for node feature reconstruction. However, their training is highly dependent on the input graph. Consequently, if the original graph is poor quality, the learned representation through multilayer aggregation becomes indiscriminate.  
 
 To improve the discriminability of representation, some recent methods learn node-specific information by considering the distinctness of each node \cite{DyFSS,dai2021poso}. NDLS \cite{zhang2021node} assigns different filtering orders to each node based on its influence score. DyFSS \cite{DyFSS} learns multiple self-supervised learning task weights derived from a gating network considering the difference in the node neighbors. In an orthogonal way, in this paper, we improve the representation learning from a feature perspective. We perform an in-depth analysis of how the ``one node one model" paradigm benefits clustering. 


\section{Methodology}
\subsection{Notation}
Define the graph data as $\mathcal{G}=\lbrace \mathcal{V},E,X\rbrace$, where $\mathcal{V}$ represents a set of $N$ nodes and $e_{ij}\in E$ denotes the edge between node $i$ and node $j$. $X=\lbrace X_1,...,X_N\rbrace^{\top}\in \mathbb{R}^{N\times d}$ is the feature matrix with $d$ dimensions, $X_i$ and $X_{.j}$ indicate the $i$-th row and the $j$-th column and of $X$, respectively. Adjacency matrix $\widetilde{A} \in \mathbb{R}^{N\times N}$ represents the graph structure. $D$ represents the degree matrix. The normalized adjacency matrix is $A = D^{-\frac{1}{2}}(\widetilde{A} + I)D^{-\frac{1}{2}}$, and the corresponding graph Laplacian is $L = I - A$.

\subsection{The Pipeline of FPGC}
The FPGC framework is shown in Fig. \ref{all2}. Our design comprises two critical components. The first component concretizes ``one node one model” with GNNs, with a squeeze-and-excitation block selecting cluster-relevant features. The second is the contrastive learning framework with a new data augmentation technique based on feature cross.

\subsection{Squeeze-and-Excitation Block}
Cluster-relevant features characterize the clusters. We design a squeeze-and-excitation block to select them. 

\textbf{Squeeze:} The squeeze operation compresses the node features from $X\in \mathbb{R}^{N \times d}$ to $q \in \mathbb{R}^{1 \times d}$. Specifically, it computes the summation of the features as follows:
\begin{equation}
q=F_{s q}\left(X\right)=\frac{1}{N} \sum_{i=1}^{N} X_i,
\end{equation}
where $q$ serves as a channel-wise statistic that captures global feature information.

\textbf{Excitation:} The excitation operation follows the squeeze step and aims to capture dependencies among all channels. It is a simple gating mechanism that uses the sigmoid function $\sigma(\cdot)$ to activate the squeezed feature map, enabling the learning of nonlinear connections between channels. Firstly, the dimension is reduced using a multi-layer perceptron (MLP) $W_1$. Next, a ReLU function $\delta(\cdot)$ and another MLP $W_2$ are used to increase the dimensionality, returning it to the original dimension $d$. Finally, the sigmoid function is applied. The process is summarized as follows:

\begin{equation}
\begin{aligned}
\tilde{q} & =F_{e x}\left(q\right) =\sigma\left(W_2 \delta\left(W_1 q\right)\right).
\end{aligned}
\end{equation}

\textbf{Selection:} The outcome of the excitation operation is considered to be the significance of each feature. Then, we define the function $F_{top}$ to select top $n$ important features according to $\tilde{q}$. This operation is defined as follows:
 \begin{equation}
X^n=F_{top}\left(\tilde{q}X\right),
\end{equation}
where $X^n \in \mathbb{R}^{N\times n}$ is the node feature matrix containing the top $n$ significant features.

\subsection{One Node One Model}
Ideally, an exclusive model should be constructed for each individual node, which is the core of our method:
\begin{equation}
    Y_i = f^{(i)}(X_i),
\end{equation}
where $Y$ and $f$ are the predicted label and the model. In this paradigm, personalization is fully maintained in the models. Unfortunately, the enormous number of nodes makes this scheme unfeasible. One possible solution is establishing several individual models for each type of cluster (the nodes in a group share personalized features). Each node can be considered as a combination of different clusters. For example, in a social network, a person is a member of both a mathematics group and several sports interest groups. Due to the diversity of these different communities, he may exhibit different characteristics when interacting with members from various communities. Specifically, the information about the mathematics group may be related to his professional research, while the information about sports clubs may be associated with his hobbies. As a result, we can decompose the output for a specific node as a combination of predictions for clusters:
\begin{equation}
Y_i=\sum_{j=1}^M w_j f^{(j)}(X_i),
\label{eq1}
\end{equation}
where $j$ denotes the model index and there are $M$ models. For an unsupervised task, learning $w_j$ directly is difficult. Instead, we generate $w_j$ from personalized features: $w_j = \left[g(X^n)\right]_j$, where $g(X^n) = \sigma\left(W_3X^n\right)$. In other words, we use personalized features to identify clusters. A higher value of a personalized feature corresponds to a higher weight. $W_3$ represents an MLP.

For simplicity, we only consider SGC \cite{SGC} as our basic model, i.e. $f^{(j)}(X_i) = A_i^kXW^j$, where $A^k$ denotes the stacked $k$-layer graph filter. Other GNNs can also be the base models. For instance, DAGNN \cite{liu2020towards} has $f^{(j)}(X_i) = \sum\limits_{t=0}^k s_t A_i^t XW^j$, where $s_t$ is the learnable weight; APPNP \cite{APPNP} has $f^{(j)}(X_i)=\operatorname{softmax}\left(\eta(\boldsymbol{I}-(1-\eta) A)^{-1}_i X\right) W^j$, where $\eta$ is the a hyper-parameter. For convenience, we define $\bar{X}=A^kX=\lbrace \bar{X}_1,...,\bar{X}_N\rbrace^{\top}$. Note that $\bar{X}_j \in \mathbb{R}^{d}$ and $W^j \in \mathbb{R}^{d \times d_{out}}$, where $d_{out}$ is the output dimension. We have:
\begin{equation}
Y_i=\sum_{j=1}^M\left[g\left(X^n\right)\right]_j \bar{X}_iW^j.
\end{equation}
The $u$-th entry of $Y_i$ is given by:
\begin{equation}
Y_{iu}=\sum_{j=1}^M \sum_{v=1}^{d}\left[g\left(X^{n}\right)\right]_j W_{uv}^j \bar{X}_{iv},
\end{equation}
which introduces a complexity of $M$ times.
Thus far, we still need to design $M$ individual models to identify clusters, which is computationally demanding. Fortunately, we have enough free parameters to simplify the process. Here we present a simple yet effective way. We set $M=d_{out}$, which gives $Y_i=\sum\limits_{j=1}^{d_{out}} \sum\limits_{v=1}^{d_{in}}\left(\left[g\left(X^n\right)\right]_j W_{uv}^j \bar{X}_{iv}\right)$. Then we let $W_{uv}^j=\left\{\begin{array}{l}W_{uv}, j=u \\ 0, j \neq u\end{array}\right.$, which yields:
\begin{equation}
\begin{aligned}
    Y_{iu}&=\left[g\left(X^{n}\right)\right]_u \sum_{v=1}^{d} W_{uv} \bar{X}_{iv} \\
    &=\left[g\left(X^{n}\right)\right]_u \bar{X}_i W_{.u},
\end{aligned}
\end{equation}
or equivalently,
\begin{equation}
Y =g\left(X^n\right) \odot \bar{X} W,
\label{view1}
\end{equation}
where $\odot$ denotes the Hadamard product. Consequently, learning a model for each node is achieved through element-wise multiplication, which is computationally efficient.

$\bold{Flexibility}$: Note that the above approach is a generic method to customize existing techniques rather than a stand-alone way. 
It can be seamlessly incorporated with many SOTA GNN models to enhance performance.

\subsection{Theoretical Analysis}
Most theoretical analysis in the GNN area focuses on graph structures \cite{mao2024demystifying}. In this section, we establish a theoretical analysis from the feature perspective. Without loss of generality, we consider the case with two clusters, $c_1$ and $c_2$. Assume the original node features follow the Gaussian distribution: $X_i \sim N\left(\boldsymbol{\mu}_1, \mathbf{I}\right) \text { for } i \in c_1\text{ and } X_i \sim N\left(\boldsymbol{\mu}_2, \mathbf{I}\right) \text { for } i \in c_2$ ($\boldsymbol{\mu}_1\boldsymbol{\mu}_2 \geq $ 0). We define the filtered feature as $\bar{X} = D^{-1}\widetilde{A}X$. We define $f_i$ and $f_j$ as models for $X_i$ and $X_j$, respectively, focusing on the personalized features in sets $T_i$ and $T_j$, while other irrelevant features are ignored. We then present the following theorem:


\begin{theorem}
Assuming the distribution of filtered features $\bar{X}$ shares the same variance $\sigma\mathbf{I}$ and the cluster has a balance distribution $\mathbf{P}\left(Y=c_1\right) = \mathbf{P}\left(Y=c_2\right)$. The upper bound of $\left|\mathbf{P}\left(Y_i=c_1 \mid f_i(\bar{X}_i)\right)-\mathbf{P}\left(Y_j=c_1 \mid f_j(\bar{X}_j)\right)\right|$ is decreasing with respect to $\sum_{u \in {T_i \cap T_j}} \bar{X}_{i u}\bar{X}_{j u}$.
\label{theo}
\end{theorem}

Note that the assumptions are not strictly necessary but are used to simplify the proof in the Appendix.
Theorem \ref{theo} indicates that if two nodes share more cluster-relevant features, they are more likely to be classified into one cluster, and vice versa.

\subsection{Contrastive Clustering }

\subsubsection{Feature Cross Augmentation}
Although graph filtering can smooth the features between node neighborhoods, it only captures high-order feature interactions and suffers
from overlooking low-order feature interaction \cite{GCNCross}. Feature cross can provide valuable information for downstream tasks. For example, features like \{$title$ \& $approach$ \& $experiment$\} can provide additional information about the paper's category. Features that collectively represent $title$ can be viewed as a field, like $title=\left\{0,1,0\right\}$. To this end, we randomly divide the filtered features into $m$ fields:
\begin{equation}
\bar{X}=\lbrace \bar{X}_{field_1},\bar{X}_{field_2},...,\bar{X}_{field_m}\rbrace.
\end{equation}
Then, every node has $m$ fields, i.e., $\bar{X_i}=\lbrace \bar{X}_{ifield_1},\bar{X}_{ifield_2},...,\bar{X}_{ifield_m}\rbrace$. We use $\bar{X}$ instead of $X$ because $\bar{X}$ has more information. The feature cross is defined as:
\begin{equation}
R_i=\left\{\bar{X}_{ifield_z} \bar{X}^\top_{ifield_j} \mid 1 \leq z<j \leq m\right\},
\end{equation}

We input the original features $X$ as a view for contrastive learning. It's well known that matrix factorization techniques can capture the low-order interaction \cite{DeepFM}. Thus, we propose to perform data augmentation as follows:
\begin{equation}
    X^{aug} = X + RW,
    \label{aug}
\end{equation}
where $W$ is an MLP to adjust the dimension. Unlike previous methods, Eq.(\ref{aug}) has two advantages. First, instead of using typical graph enhancements such as feature masking or edge perturbation, our method generates augmented views from filtered features. This maintains the semantics of the augmented view. Second, other learnable augmentations design complex losses to remain close to the initial features. Eq.(\ref{aug}) is based on the feature cross, which can provide rich semantic information. The complexity of the feature cross is $O(Nm^2)$, which is linear to $N$. The results can be stored and accessed after a one-time computation. 

Similarly to Eq.(\ref{view1}), the second view is finally formulated as:
\begin{equation}
Y^{\prime}=g\left(X^n\right) \odot X^{aug} W.
\label{view2}
\end{equation}
Note that the same node in two different views shares the same model.

\subsubsection{Loss Function}
For two views $Y$ and $Y^{\prime}$, we treat the same node in different views as positive samples and all other nodes as negative samples. The pairwise loss is defined as follows:
\begin{equation}
\begin{aligned}
&\ell\left(Y_i, Y^{\prime}_i\right)=-\log \frac{e^{\operatorname{sim}\left(Y_i, Y^{\prime}_i\right)  }}{\sum_{j=1}^N e^{\operatorname{sim}\left(Y_i, Y^{\prime}_j\right) }+\sum_{j=1}^N e^{\operatorname{sim}\left(Y_i, Y_j\right)}},\\
&\mathcal{L}_{con}=\frac{1}{2 N} \sum_{i=1}^N\left[\ell\left(Y_i, Y^{\prime}_i\right)+\ell\left(Y^{\prime}_i, Y_i\right)\right],
\end{aligned}
\end{equation}
where $\operatorname{sim}$ is the cosine similarity. Besides, the reconstruction loss can be calculated as follows:
\begin{equation}
\mathcal{L}_{re}=\frac{1}{N^2}\left\| Y^{\prime}Y^\top - A\right\|_F.
\end{equation}

Finally, the total loss is formulated as:
\begin{equation}
\mathcal{L}= \mathcal{L}_{re}+\lambda\mathcal{L}_{con},
\end{equation}
where $\lambda>0$ is a trade-off parameter. The clustering results are obtained by performing K-means on $\frac{1}{2}(Y+Y^{\prime})$. The detailed learning process of FPGC is illustrated in the Appendix.

\section{Experiments}


\subsection{Datasets}
We select seven graph clustering benchmark datasets, which are: Cora \cite{Cora}, CiteSeer \cite{Cora}, Pubmed \cite{Cora}, Amazon Photo (AMAP) \cite{AMAP}, USA Air-Traffic (UAT) \cite{EAT}, Europe Air-Traffic (EAT) \cite{EAT}, and Brazil Air Traffic (BAT) \cite{EAT}. We also add two large-scale graph datasets, the image relationship network Flickr \cite{Flickr} and the social network Twitch-Gamers \cite{twitch}. To see the relevance between graph structure and downstream task, we also report the Aggregation Class Distance (ACD) \cite{BMGC}.
The statistics information is summarized in Table \ref{tab::datasets}. We can see that these datasets are inherently low-quality.

 \begin{table}[t]
		\centering

	\caption{Statistics information of datasets.}
		\label{tab::datasets}
            \resizebox{.4\textwidth}{!}{
    \begin{tabular}{cccccccc}
    \toprule
    \multicolumn{2}{c}{Graph datasets} & Nodes & Dims. & Edges & Clusters & ACD\\
    \midrule
        & Cora  & 2708  & 1433  & 5429  & 7  &0.52 \\
          & Citeseer & 3327  & 3703  & 4732  &6 &0.36    \\
          & Pubmed & 19717 &500 &44327 &3 &0.36 \\
          & UAT   & 1190  & 239  & 13599 & 4  &0.35   \\
          & AMAP  & 7650  & 745   & 119081 & 8 &0.28 \\
          & EAT   & 399   & 203   & 5994  & 4 &0.25 \\
          & BAT   & 1190  & 239   & 13599 & 4  & 0.46\\
          & Flickr &89250 &500 &899756 & 7 &0.02 \\
          & Twitch-Gamers &168114 &7 &67997557 &2 &0.09\\
    \bottomrule
    \end{tabular}
}
	\end{table}

 \subsection{Comparison Methods}
To demonstrate the superiority of FPGC, we compare it to several recent baselines. These methods can be roughly divided into three kinds: 1) traditional GNN-based methods: SSGC \cite{SSGC}. 2) shallow methods: MCGC \cite{MCGC}, FGC \cite{FGC}, and CGC \cite{CGC}. 
3) contrastive learning-based methods: MVGRL \cite{MVGRL}, SDCN \cite{SDCN}, DFCN \cite{DFCN}, SCGC \cite{SCGC}, CCGC \cite{CCGC}, and CONVERT \cite{convert}. 
4) Advanced autoencoder-based methods: AGE \cite{AGE}, DGCN \cite{DGCN}, DMGNC \cite{DMGNC}, and DyFSS \cite{DyFSS}.

 \subsection{Experimental Setting}
To ensure fairness, all experimental settings follow the DGCN \cite{DGCN}, which performs a grid search to find the best results. Our network is trained with the Adam optimizer for 400 epochs until convergence. Three MLPs consist of a single embedding layer, with 100 dimensions on Flickr and Twitch-Gamers, and 500 dimensions on other datasets. The learning rate is set to 1e-2 on BAT/EAT/UAT/Twitch-Gamers, 1e-3 on Cora/Citeseer/Pumbed/Flickr, and 1e-4 on AMAP. The graph aggregating layers $k$ is searched in \{2,3,4,5\}. The number of fields $m$ is set according to the density of features, i.e., denser features should have smaller values to cross fewer times, the number of important features $n$ is set according to the number of features, i.e., more features should have larger values, the trade-off parameter $\lambda$ is tuned in \{0.001, 0.1, 1, 100\}. Thus, $\left\{k, m, n, \lambda\right\}$ are set to $\left\{3, 60, 100, 1\right\}$ on Cora, $\left\{4, 50, 50, 0.1\right\}$ on Citeseer, $\left\{5, 10, 10, 0.001\right\}$ on Pubmed, $\left\{5, 20, 10, 100\right\}$ on UAT, $\left\{2, 10, 10, 0.001\right\}$ on AMAP, $\left\{4, 10, 10, 0.001\right\}$ on EAT, $\left\{5, 20, 20, 1\right\}$ on BAT, $\left\{4, 10, 100, 1\right\}$ on Flickr and $\left\{2, 2, 4, 1\right\}$ on Twitch-Gamers. We evaluate clustering performance with two widely used metrics: ACC and NMI. All experiments are conducted on the same machine with the Intel(R) Core(TM) i9-12900k CPU, two GeForce GTX 3090 GPUs, and 128GB RAM \footnote{The code is available at https://github.com/XieXuanting/FPGC}.

\begin{table*}[!htbp]
		\centering
  \caption{Clustering Results. The best performance is marked in  \textcolor[rgb]{ 1,  0,  0}{\textbf{red}}. ``-" indicates the original paper does not have this result and the provided code can't produce the result.}
		\label{Rehomo}%
  \resizebox{0.8\textwidth}{!}{
        \begin{tabular}{ccccccccccccccc}
    \toprule
    \multirow{2}[4]{*}{Methods} & \multicolumn{2}{c}{Cora} & \multicolumn{2}{c}{Citeseer}  & \multicolumn{2}{c}{Pubmed} & \multicolumn{2}{c}{UAT} & \multicolumn{2}{c}{AMAP} & \multicolumn{2}{c}{EAT} & \multicolumn{2}{c}{BAT} \\
\cmidrule{2-15}          & ACC   & NMI   & ACC   & NMI   & ACC   & NMI   & {ACC} & NMI   & ACC   & NMI   & ACC   & NMI  & ACC   & NMI \\
    \midrule
    DFCN  & 36.33  & 19.36 & 69.50  & 43.90  &-&-& 33.61  & 26.49  & 76.88 & 69.21  & 32.56 & 8.27 & 35.56 & 8.25\\

    SSGC  & 69.60  & 54.71 & 69.11  & 42.87 &-&- & 36.74  & 8.04 & {60.23} & 60.37 &32.41 &4.65 & 36.74 &8.04 \\
    MVGRL & 70.47  & 55.57 & 68.66  & 43.66 &-&-& 44.16  & 21.53 & {45.19} & 36.89 &32.88 &11.72 &37.56 &29.33 \\
    SDCN  & 60.24  & 50.04 & 65.96  & 38.71 &65.78&29.47& 52.25  & 21.61 & {53.44} & 44.85  & 39.07 &8.83 &53.05 &25.74 \\
    AGE   & 73.50  & 57.58 & 70.39  & 44.92 &-&-& 52.37  & 23.64 & {75.98} & -    & 47.26 & 23.74 & 56.68 & 36.04  \\
    MCGC  & 42.85  & 24.11 & 64.76  & 39.11 &66.95&32.45& 41.93  & 16.64 & {71.64} & 61.54   & 32.58 & 7.04 &38.93 & 23.11\\
    FGC   & 72.90  & 56.12 & 69.01  & 44.02&70.01&31.56 & 53.03  & 27.06 & {71.04} & -      & 36.84 & 10.07 & 47.33 & 18.90\\
    CONVERT &74.07&55.57 &68.43&41.62 &68.78&29.72 &57.36&28.75 &77.19&62.70 &58.35&33.36 &78.02&53.54 \\
    SCGC & 73.88 & 56.10 &71.02 &45.25 &67.73&28.65&56.58 &28.07 &77.48 &67.67   & 57.94 & 33.91 & 77.97 & 52.91 \\
    CGC & 75.15 & 56.90 &69.31 &43.61 &67.43&33.07&49.58&17.49 &73.02&63.26 &44.32 & 30.25 &53.44 & 26.97 \\
    CCGC & 73.88 & 56.45 & 69.84 & 44.33 &68.06&30.92&56.34 &28.15 &77.25 & 67.44   & 57.19 & 33.85 &75.04 & 50.23 \\
    DGCN & 72.19 & 56.04 & 71.27 & 44.13 &-&-& 52.27 & 23.54 & 76.07 & 66.13  &51.27 &31.98 &70.15 &49.52 \\
    DMGNC & 73.12 & 54.80 & 71.27 & 44.40 &70.46&34.21&- &- &- &-  &- &- &- &- \\
    DyFSS &72.19 &55.49 &70.18 &44.80 &68.05 &26.87 &51.43 &25.52 &76.86 &67.78 &43.36 &21.23 &77.25 &51.33\\
    FPGC & \textcolor[rgb]{ 1,  0,  0}{\textbf{79.19}} & \textcolor[rgb]{ 1,  0,  0}{\textbf{59.55}} & \textcolor[rgb]{ 1,  0,  0}{\textbf{72.59}} & \textcolor[rgb]{ 1,  0,  0}{\textbf{46.36}} &\textcolor[rgb]{ 1,  0,  0}{\textbf{71.03}} &\textcolor[rgb]{ 1,  0,  0}{\textbf{34.57}} & \textcolor[rgb]{ 1,  0,  0}{\textbf{58.07}} &\textcolor[rgb]{ 1,  0,  0}{\textbf{30.64}} & \textcolor[rgb]{ 1,  0,  0}{\textbf{78.44}} & \textcolor[rgb]{ 1,  0,  0}{\textbf{69.40}} &\textcolor[rgb]{ 1,  0,  0}{\textbf{58.90}} &\textcolor[rgb]{ 1,  0,  0}{\textbf{34.24}} &\textcolor[rgb]{ 1,  0,  0}{\textbf{79.38}} &\textcolor[rgb]{ 1,  0,  0}{\textbf{55.57}} \\
    \bottomrule
    \label{resultall}
    \end{tabular}}%
	\end{table*}%

\subsection{Results Analysis}
The results are illustrated in Table \ref{resultall}. We find that FPGC achieves dominant performance in all cases. For example, on the Cora dataset, FPGC surpasses the runner-up by 4.04\% and 2.65\% in terms of ACC and NMI. Traditional GNN-based methods have poor performance compared to other methods, which dig for more structure information or use contrastive learning to implicitly capture the supervision information. Note that SCGC, CCGC, and CONVERT are the most recent contrastive methods that design new augmentations. Our method constantly exceeds advanced GNN-based methods with adaptive filters, which indicates the significance of fully exploring feature information. 

DMGNC and DyFSS are the latest methods that explicitly consider feature information. DMGNC's performance does not show clear advantages compared to other baselines on Cora and Citeseer. Though applying a node-wise feature fusion strategy, DyFSS also performs poorly. For example, our method's ACC and NMI are 7.00\% and 4.06\% higher on Cora, and 15.54\% and 13.01\% higher on EAT, respectively. This is because they perform general embedding without exploiting the relationship between features and clusters.


\subsection{Parameter Analysis}
\begin{figure}[t]
    \centering
    \includegraphics[width=1.\linewidth]{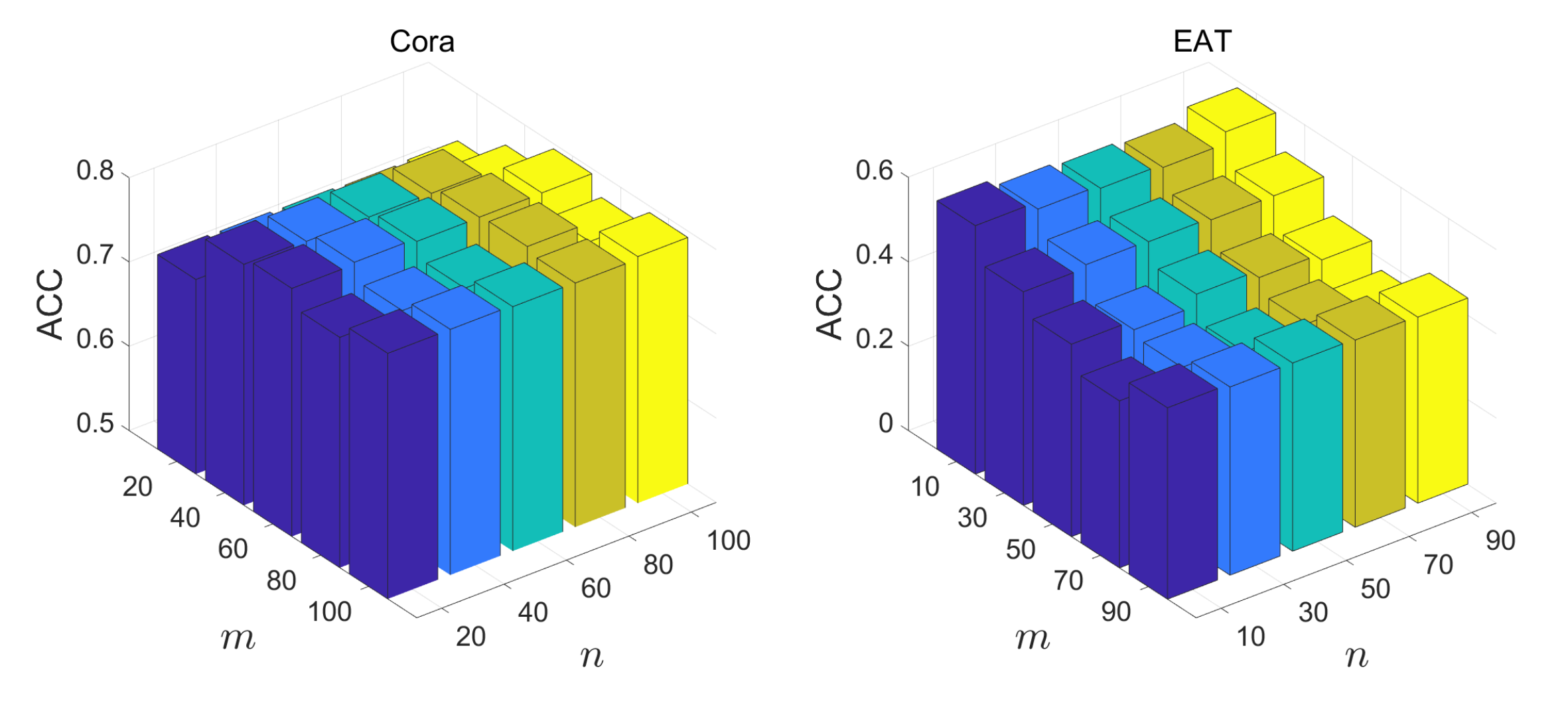}
    \caption{Parameter analysis of $m$ and $n$ on Cora and EAT.}
    \label{fig2}
\end{figure}

 To assess the impact of parameters, we evaluate the clustering accuracy of FPGC across them on Cora and EAT.
First, we test the performance with different $m$ and $n$. As shown in Fig. \ref{fig2}, Cora is less sensitive to these two parameters than EAT. In addition, Cora prefers a large $m$ while EAT opts for a small $m$, this is consistent with the density (proportion of non-zero values) of $\bar{X}$: 38.69\% on Cora and 50.25\% on EAT. Too large $m$ will degrade the performance since excessive augmentation could deteriorate the original information.  Cora prefers a large $n$ while EAT opts for a small one, which is reasonable since the attribute dimension of Cora is much larger.

\begin{figure}[t]
    \centering
    \includegraphics[width=1.\linewidth]{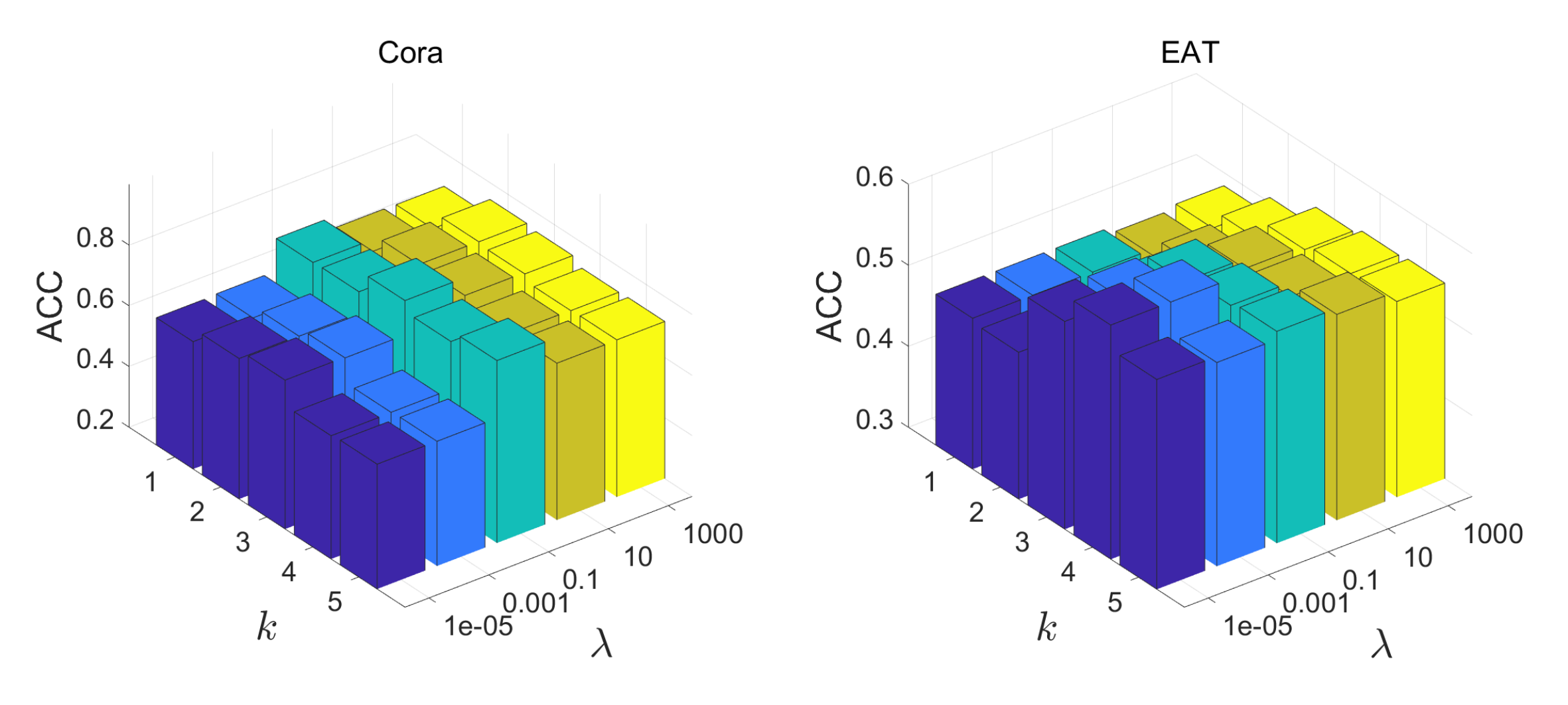}
    \caption{Parameter analysis of $k$ and $\lambda$ on Cora and EAT.}
    \label{fig3}
\end{figure}

Secondly, the impact of $k$ and $\lambda$ is shown in Fig. \ref{fig3}. FPGC can perform in effect for a wide range of $\lambda$. A small $k$ is enough to achieve a decent result.

\subsection{Results on Large-scale Data }
\begin{figure}[t]
    \centering
    \includegraphics[width=.49\linewidth]{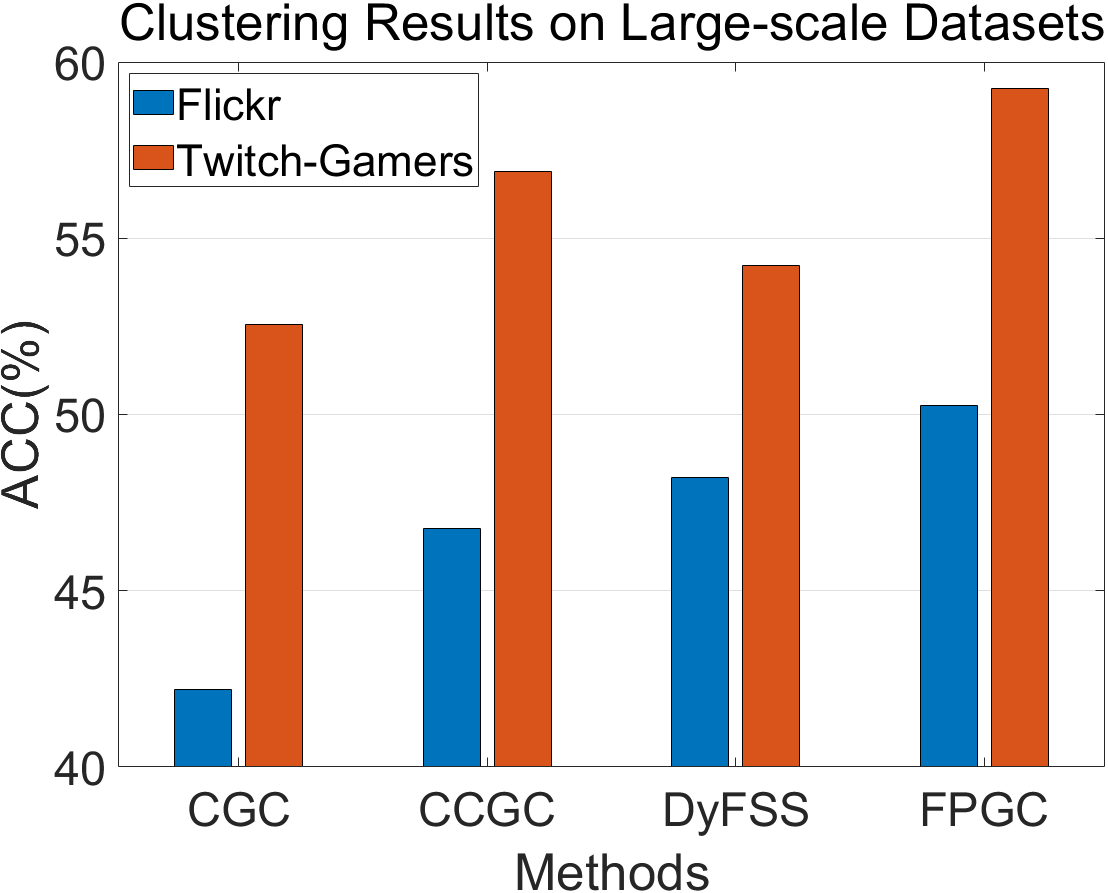}
    \includegraphics[width=.49\linewidth]{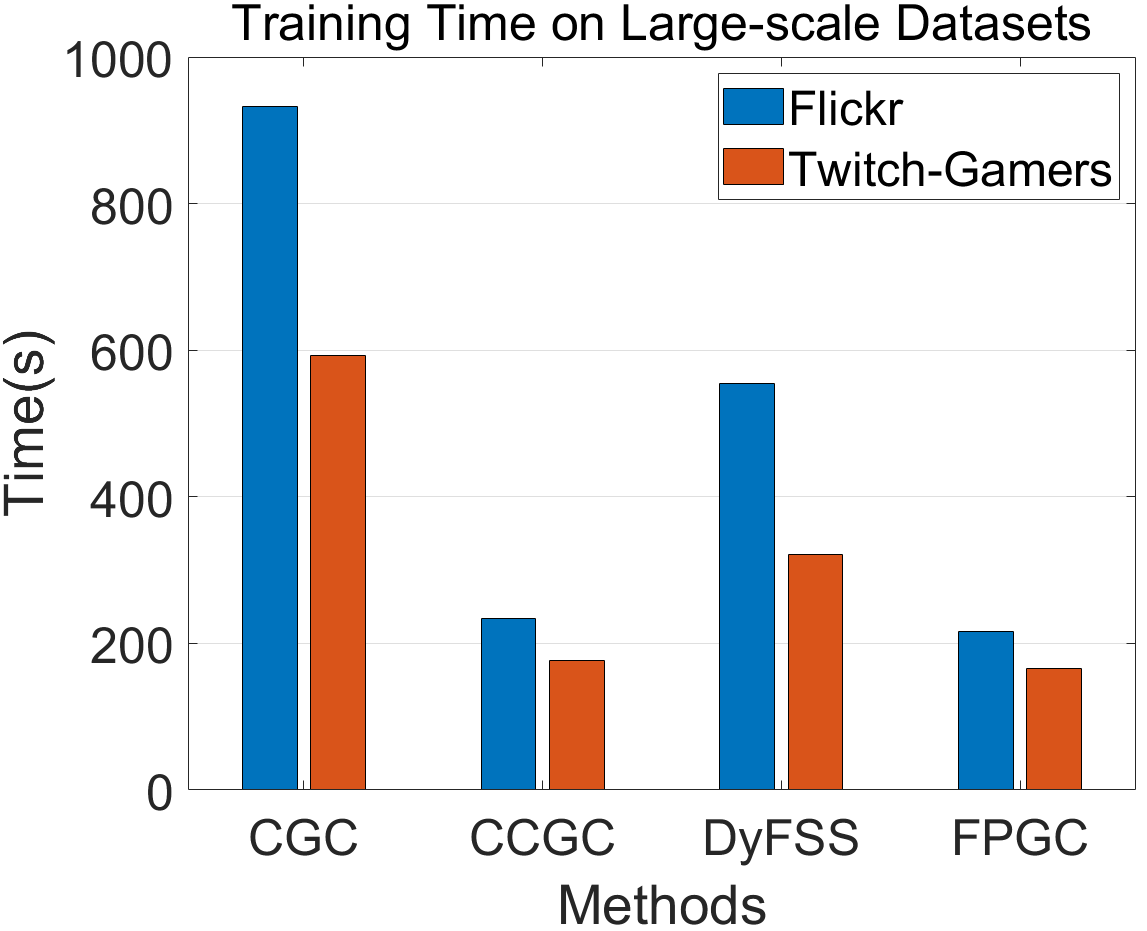}    
    \caption{Clustering results and running time on Flickr and Twitch-Gamers.}
    \label{fig45}
\end{figure}

To evaluate the scalability of FPGC, we conduct the experiments on large graph datasets Flickr and Twitch-Gamers, which have 89250 and 168114 nodes, respectively. Note that many methods fail to run on these datasets, such as DGCN. We set the batch size to 1000 for all methods. We select three recent baselines to see their performance and the total training time. Fig. \ref{fig45} shows the results. FPGC continues to outperform all comparison approaches. The required training time of FPGC is notably lower than that of the CGC and DyFSS and exhibits a marginal advantage over CCGC. This is because the feature-personalized step involves only matrix multiplication, while graph filtering and feature crossing can be pre-computed and stored. The result not only highlights the effectiveness of FPGC but also underscores its scalability and efficiency in handling large-scale datasets.

\begin{figure}[t]
    \centering
    \includegraphics[width=.9\linewidth]{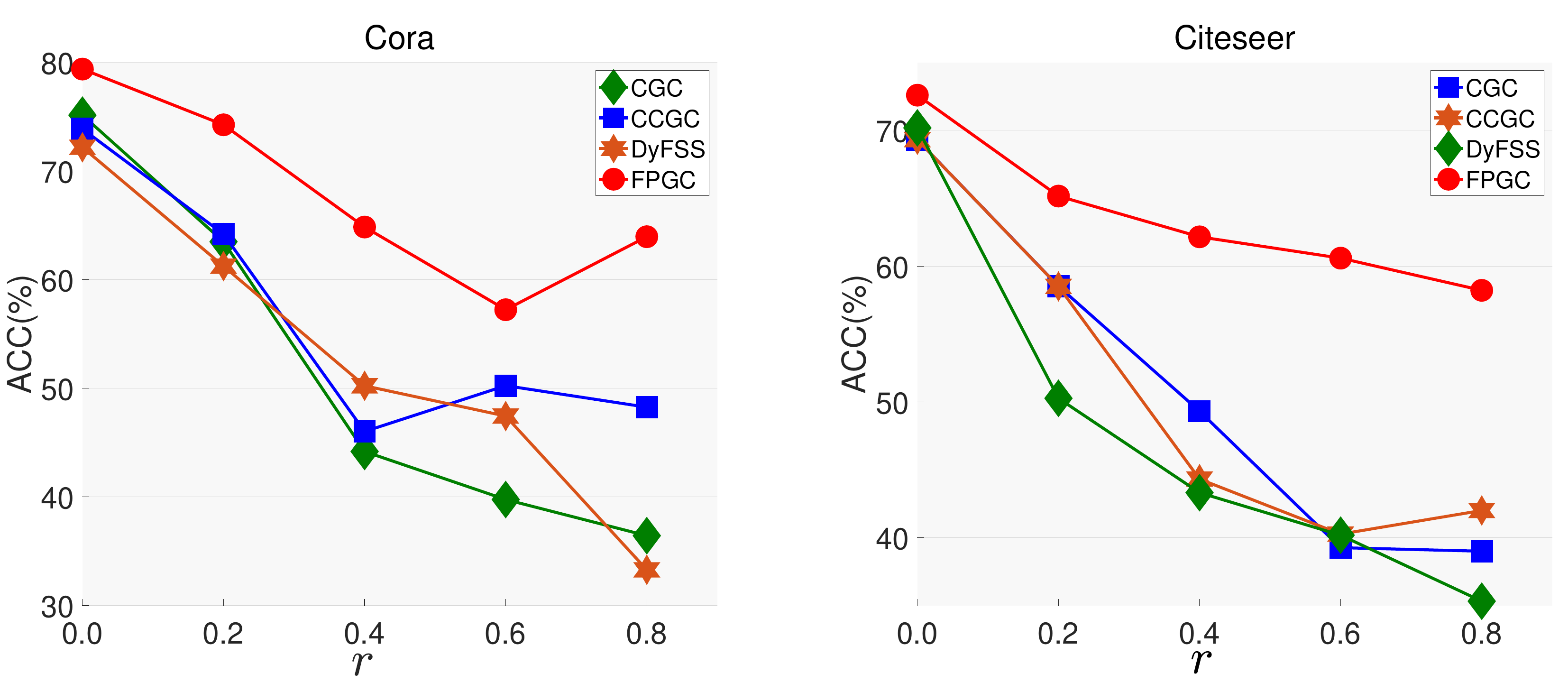}    
    \caption{Results of robustness test on Cora (left) and Citeseer (right).}
    \label{fig7}
\end{figure}

\begin{table*}[!htbp]
\centering
\caption{Results of ablation study. The best performance is marked in \textbf{\color{red}{red}} and the runner-up is marked in \textbf{bold}.}\label{abla1}
\begin{center}
\resizebox{.73\textwidth}{!}{
\begin{tabular}{ l rl| rl| rl| rl | rl}

\midrule
Methods & \multicolumn{2}{c}{{FPGC $w/o$ $aug$}} & \multicolumn{2}{c}{{FPGC $w$ $aug$}} & \multicolumn{2}{c}{{FPGC $w/o$ $g()$}} & \multicolumn{2}{c}{{FPGC $w$ DAGNN}} & \multicolumn{2}{c}{{FPGC}} \\ 
\cmidrule(lr){2-3} \cmidrule(lr){4-5} \cmidrule(lr){6-7} \cmidrule(lr){8-9} \cmidrule(lr){10-11} 
    & {ACC} & {NMI} & {ACC} & {NMI}& {ACC} & {NMI} & {ACC} & {NMI} & {ACC} & {NMI} \\
\midrule

Cora

&75.95 &58.98
&76.42 &57.52
&\textbf{78.52}&58.72
&78.27 &\textbf{59.02}
&\textcolor{red}{\textbf{79.19}} &\textcolor{red}{\textbf{59.55}}

\\
Citeseer
&71.04 &44.90
&71.33 &44.65
&70.23&44.34
&\textbf{72.10} &\textbf{46.02}
&\textcolor{red}{\textbf{72.59}} &\textcolor{red}{\textbf{46.36}}
 \\
Pubmed
&66.98 &28.47
&67.76 &29.42
&\textbf{71.20}&\textbf{34.96}
&\textcolor{red}{\textbf{72.05}} &\textcolor{red}{\textbf{35.26}}
&71.03 &34.57
\\

UAT
&57.25 &29.06
&57.61 &29.69
&57.13&29.86
&\textcolor{red}{\textbf{58.24}} &\textcolor{red}{\textbf{30.72}}
&\textbf{58.07} &\textbf{30.64}

\\

AMAP
&77.45 &68.32
&78.01 &68.96
&76.88 &66.74
&\textcolor{red}{\textbf{79.52}} &\textcolor{red}{\textbf{70.22}}
&\textbf{78.44} &\textbf{69.40}

\\

EAT
&57.82 &33.32
&58.05 &33.96
&57.25 &33.19
&\textcolor{red}{\textbf{59.28}} &\textcolor{red}{\textbf{34.71}}
&\textbf{58.90} &\textbf{34.24}

\\

BAT
&78.12 &53.81
&78.23 &53.96
&77.23 &52.19
&\textbf{79.03} &\textbf{55.02}
&\textcolor{red}{\textbf{79.38}} &\textcolor{red}{\textbf{55.57}}

\\

\bottomrule
\end{tabular}}
\end{center}
\end{table*}

\begin{figure}[!htbp]
    \centering
    \includegraphics[width=.73\linewidth]{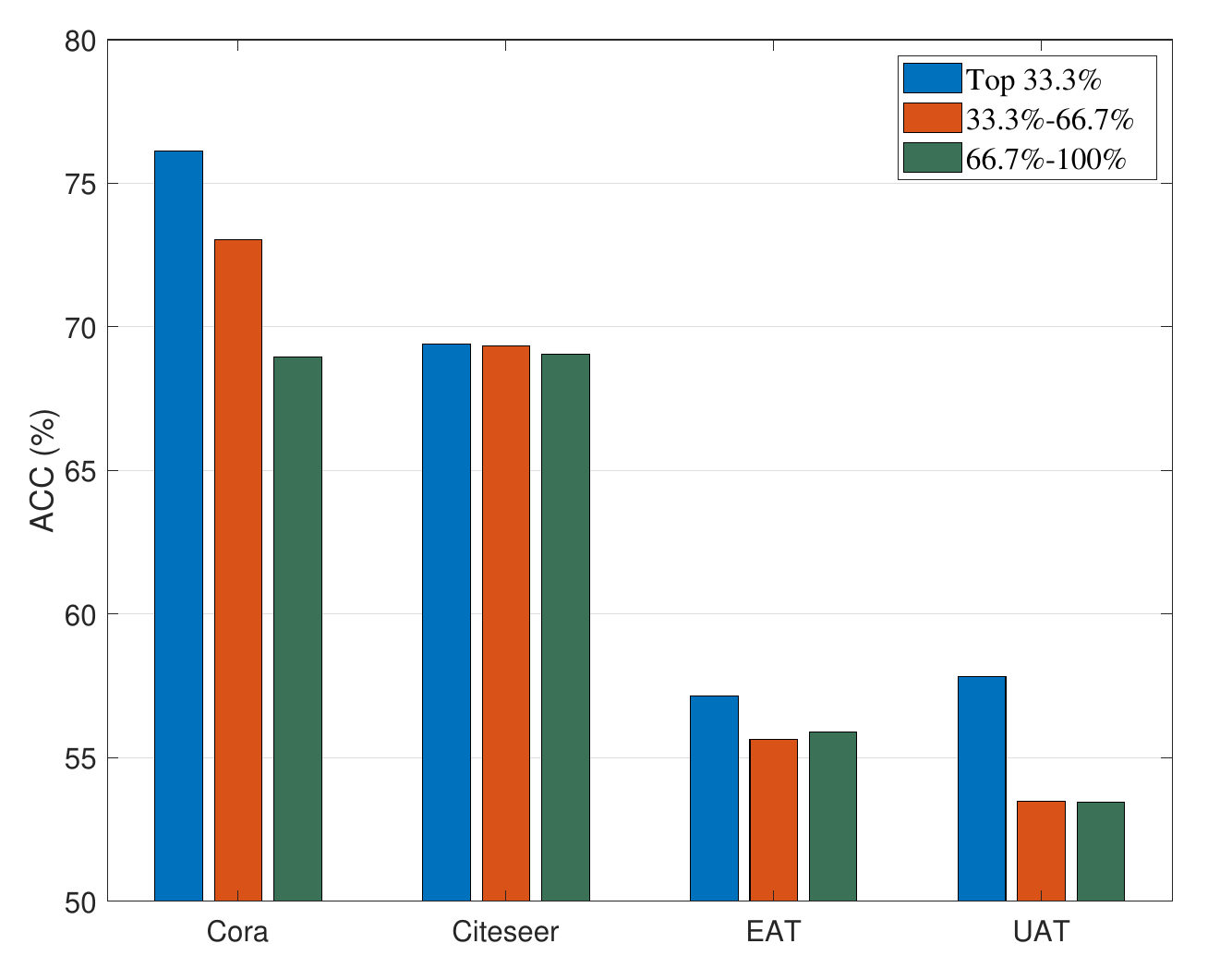}
    \caption{The results on different important features.}
    \label{sth04}
\end{figure}
\subsection{Robustness Analysis}
To demonstrate the significance of the feature, we examine the situation in which the graph structure is of poor quality. Specifically, we add edges to the graphs and remove the same number of original edges from them at random \cite{fu2022p}. We define $r = \frac{\text{\#random edges}}{\text{\#all edges}}$ as the random edge rate. With $r$ = \{0.2, 0.4, 0.6, 0.8\}, we report the ACC of CGC, CCGC, DyFSS, and FPGC in Cora and Citeseer. From Fig. \ref{fig7}, it can be seen that the PFGC achieves the best performance in all cases. Especially when the perturbation rate is extremely high, our method shows a more stable tendency than other clustering methods. The performance of other methods changes dramatically, indicating that they rely highly on the graph structure. Thus, our method is robust to the graph structure noise.

\subsection{Ablation Study}
To assess the impact of feature cross, we test the clustering performance after removing it, marking it as ``FPGC $w/o$ $aug$". The results are shown in Table \ref{abla1}. It is clear that feature cross does improve the clustering performance. In addition, we replace feature cross with classical graph augmentation strategies, including drop edges \cite{aug1}, add edges \cite{aug2}, graph diffusion \cite{MVGRL}, and mask feature \cite{yu2022sail}. We set the change rate to 20\% according to the suggestions of the original paper. The best performance among these four methods is marked as ``FPGC $w$ $aug$". Our method can still beat them, thus feature cross can provide more rich information. 

To see the influence of feature selection, we test the performance without $g(X^n)$ and mark it as ``FPGC $w/o$ $g()$". We can see the performance degradation in most cases in Table \ref{abla1}. Thus learning a model for each node can successfully keep each node's personality. FPGC does not achieve better performance on Pubmed. This may be because Pubmed has a large node size but only has 3 clusters, which means that it is difficult to select the cluster-relevant features in this case. We also test the performance with DAGNN \cite{liu2020towards} as the base model and mark this method as ``FPGC $w$ DAGNN". It can be seen that it achieves the best performance in 4 out of 7 cases in Table \ref{abla1}. Thus, our proposed ``one node one model" paradigm is promising with other SOTA GNNs.

To verify that the proposed squeeze-and-excitation block can select essential features, we categorize them into three intervals: features with the top 33.3\% (highest) $\tilde{q}$, features from 33.3\% to 66.7\%, and the remaining 66.7\% to 100\%. We use these indexes to select features for $X^n$ and test their performance, respectively. From Fig. \ref{sth04}, we can see that when selecting the most important features (top 33.3\%), we achieve the best results in all cases. This indicates that features with high value in $\tilde{q}$ are vital.  Therefore, our proposed squeeze-and-excitation block can successfully assign high weights to essential features, which is beneficial for downstream task.

\section{Conclusion}
In this paper, we introduce the ``one node one model" paradigm,  which addresses the limitations of existing graph clustering methods by emphasizing missing-half feature information. By incorporating a squeeze-and-excitation block to select cluster-relevant features, our method effectively enhances the discriminative power of the learned representations. The proposed feature cross data augmentation further enriches the information available for contrastive learning. Our theoretical analysis and experimental results demonstrate that our method significantly improves clustering performance. 

\section*{Acknowledgments}
This work was supported by the National Natural Science Foundation of China (No. U24A20323).

\bibliography{aaai25}

\clearpage  
\includepdf[pages=1]{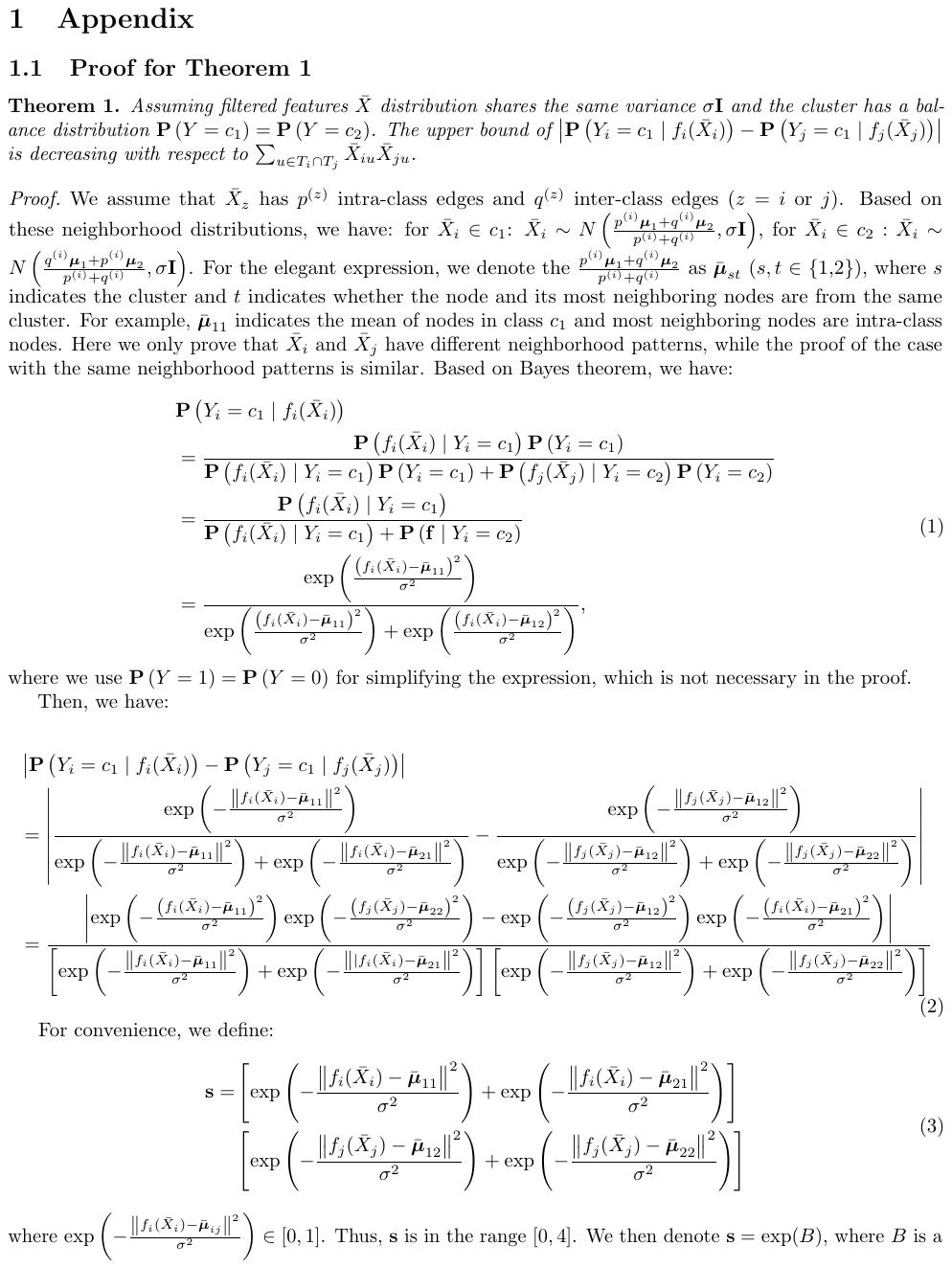}
\includepdf[pages=2]{Appendix.pdf}
\includepdf[pages=3]{Appendix.pdf}

\end{document}


\section{Appendix}
\subsection{Proof for Theorem 1}
\begin{theorem}
 Assuming filtered features $\bar{X}$ distribution shares the same variance $\sigma \mathbf{I}$ and the cluster has a balance distribution $\mathbf{P}\left(Y=c_1\right) = \mathbf{P}\left(Y=c_2\right)$. The upper bound of $\left|\mathbf{P}\left(Y_i=c_1 \mid f_i(\bar{X}_i)\right)-\mathbf{P}\left(Y_j=c_1 \mid f_j(\bar{X}_j)\right)\right|$ is decreasing with respect to $\sum_{u \in {T_i \cap T_j}} \bar{X}_{i u}\bar{X}_{j u}$.
\label{theo}
\end{theorem}

\begin{proof}
We assume that $\bar{X}_z$ has $p^{(z)}$ intra-class edges and $q^{(z)}$ inter-class edges ($z$ = $i$ or $j$). Based on these neighborhood distributions, we have: for $\bar{X}_i \in c_1$: $\bar{X}_i \sim N\left(\frac{p^{(i)}\boldsymbol{\mu}_1+q^{(i)} \boldsymbol{\mu}_2}{p^{(i)}+q^{(i)}}, \sigma \mathbf{I}\right)$, for $\bar{X}_i \in c_2$ $ : \bar{X}_i \sim N\left(\frac{q^{(i)} \boldsymbol{\mu}_1+p^{(i)} \boldsymbol{\mu}_2}{p^{(i)}+q^{(i)}}, \sigma \mathbf{I}\right)$. For the elegant expression, we denote the $\frac{p^{(i)} \boldsymbol{\mu}_1+q^{(i)} \boldsymbol{\mu}_2}{p^{(i)}+q^{(i)}}$ as $\bar{\boldsymbol{\mu}}_{st}$ ($s,t$ $\in$ \{1,2\}), where $s$ indicates the cluster and $t$ indicates whether the node and its most neighboring nodes are from the same cluster. For example, $\bar{\boldsymbol{\mu}}_{11}$ indicates the mean of nodes in class $c_1$ and most neighboring nodes are intra-class nodes. Here we only prove that $\bar{X}_i$ and $\bar{X}_j$ have different neighborhood patterns, while the proof of the case with the same neighborhood patterns is similar. Based on Bayes theorem, we have:

\begin{equation}
\begin{aligned}
& \mathbf{P}\left(Y_i=c_1 \mid f_i(\bar{X}_i)\right) \\
& =\frac{\mathbf{P}\left(f_i(\bar{X}_i) \mid Y_i=c_1\right) \mathbf{P}\left(Y_i=c_1\right)}{\mathbf{P}\left(f_i(\bar{X}_i) \mid Y_i=c_1\right) \mathbf{P}\left(Y_i=c_1\right)+\mathbf{P}\left(f_j(\bar{X}_j) \mid Y_i=c_2\right) \mathbf{P}\left(Y_i=c_2\right)} \\
& =\frac{\mathbf{P}\left(f_i(\bar{X}_i) \mid Y_i=c_1\right)}{\mathbf{P}\left(f_i(\bar{X}_i) \mid Y_i=c_1\right)+\mathbf{P}\left(\mathbf{f} \mid Y_i=c_2\right)} \\
& =\frac{\exp \left(\frac{\left(f_i(\bar{X}_i)-\bar{\boldsymbol{\mu}}_{11}\right)^2}{\sigma^2}\right)}{\exp \left(\frac{\left(f_i(\bar{X}_i)-\bar{\boldsymbol{\mu}}_{11}\right)^2}{\sigma^2}\right)+\exp \left(\frac{\left(f_i(\bar{X}_i)-\bar{\boldsymbol{\mu}}_{12}\right)^2}{\sigma^2}\right)},
\end{aligned}
\end{equation}
where we use $\mathbf{P}\left(Y=1\right) = \mathbf{P}\left(Y=0\right)$ for simplifying the expression, which is not necessary in the proof.

Then, we have:

\begin{equation}
\begin{aligned}
& \left|\mathbf{P}\left(Y_i=c_1 \mid f_i(\bar{X}_i)\right) -\mathbf{P}\left(Y_j=c_1 \mid f_j(\bar{X}_j)\right)\right| \\
& =\left|\frac{\exp \left(-\frac{\left\|f_i(\bar{X}_i)-\bar{\boldsymbol{\mu}}_{11}\right\|^2}{\sigma^2}\right)}{\exp \left(-\frac{\left\|f_i(\bar{X}_i)-\bar{\boldsymbol{\mu}}_{11}\right\|^2}{\sigma^2}\right)+\exp \left(-\frac{\left\|f_i(\bar{X}_i)-\bar{\boldsymbol{\mu}}_{21}\right\|^2}{\sigma^2}\right)}-\frac{\exp \left(-\frac{\left\|f_j(\bar{X}_j)-\bar{\boldsymbol{\mu}}_{12}\right\|^2}{\sigma^2}\right)}{\exp \left(-\frac{\left\|f_j(\bar{X}_j)-\bar{\boldsymbol{\mu}}_{12}\right\|^2}{\sigma^2}\right)+\exp \left(-\frac{\left\|f_j(\bar{X}_j)-\bar{\boldsymbol{\mu}}_{22}\right\|^2}{\sigma^2}\right)}\right| \\
& =\frac{\left|\exp \left(-\frac{\left(f_i(\bar{X}_i)-\bar{\boldsymbol{\mu}}_{11}\right)^2}{\sigma^2}\right) \exp \left(-\frac{\left(f_j(\bar{X}_j)-\bar{\boldsymbol{\mu}}_{22}\right)^2}{\sigma^2}\right)-\exp \left(-\frac{\left(f_j(\bar{X}_j)-\bar{\boldsymbol{\mu}}_{12}\right)^2}{\sigma^2}\right) \exp \left(-\frac{\left(f_i(\bar{X}_i)-\bar{\boldsymbol{\mu}}_{21}\right)^2}{\sigma^2}\right)\right|}{\left[\exp \left(-\frac{\left\|f_i(\bar{X}_i)-\bar{\boldsymbol{\mu}}_{11}\right\|^2}{\sigma^2}\right)+\exp \left(-\frac{\left\|\mid f_i(\bar{X}_i)-\bar{\boldsymbol{\mu}}_{21}\right\|^2}{\sigma^2}\right)\right]\left[\exp \left(-\frac{\left\|f_j(\bar{X}_j)-\bar{\boldsymbol{\mu}}_{12}\right\|^2}{\sigma^2}\right)+\exp \left(-\frac{\left\|f_j(\bar{X}_j)-\bar{\boldsymbol{\mu}}_{22}\right\|^2}{\sigma^2}\right)\right]}
\end{aligned}
\end{equation}

For convenience, we define:
\begin{equation}
\begin{aligned}
\mathbf{s}= &\left[\exp \left(-\frac{\left\|f_i(\bar{X}_i)-\bar{\boldsymbol{\mu}}_{11}\right\|^2}{\sigma^2}\right)+\exp \left(-\frac{\left\|f_i(\bar{X}_i)-\bar{\boldsymbol{\mu}}_{21}\right\|^2}{\sigma^2}\right)\right]\\
&\left[\exp \left(-\frac{\left\|f_j(\bar{X}_j)- \bar{\boldsymbol{\mu}}_{12}\right\|^2}{\sigma^2}\right)+\exp \left(-\frac{\left\|f_j(\bar{X}_j)-\bar{\boldsymbol{\mu}}_{22}\right\|^2}{\sigma^2}\right)\right] \\
\end{aligned}
\end{equation}
where $\exp \left(-\frac{\left\|f_i(\bar{X}_i)-\bar{\boldsymbol{\mu}}_{i j}\right\|^2}{\sigma^2}\right) \in[0,1]$. Thus, $\mathbf{s}$ is in the range $[0,4]$. We then denote $\mathbf{s}=\exp (B)$, where $B$ is a constant, which produces:
\begin{equation}
\begin{aligned}
& \left|\mathbf{P}\left(Y_i=c_1 \mid f_i(\bar{X}_i)\right)-\mathbf{P}\left(Y_j=c_1 \mid f_j(\bar{X}_j)\right)\right| \\
& =-\frac{\left|\exp \left(-\frac{\left(f_i(\bar{X}_i)-\bar{\boldsymbol{\mu}}_{11}\right)^2}{\sigma^2}\right) \exp \left(-\frac{\left(f_j(\bar{X}_j)-\bar{\boldsymbol{\mu}}_{22}\right)^2}{\sigma^2}\right)-\exp \left(-\frac{\left(f_i(\bar{X}_i)-\bar{\boldsymbol{\mu}}_{21}\right)^2}{\sigma^2}\right) \exp \left(-\frac{\left(f_j(\bar{X}_j)-\bar{\boldsymbol{\mu}}_{12}\right)^2}{\sigma_2^2}\right)\right|}{\exp (B)} \\
& =\left|\exp \left(-\frac{\left(f_i(\bar{X}_i)-\bar{\boldsymbol{\mu}}_{11}\right)^2+\left(f_j(\bar{X}_j)-\bar{\boldsymbol{\mu}}_{22}\right)^2}{\sigma^2}+B\right)-\exp \left(-\frac{\left(f_i(\bar{X}_i)-\bar{\boldsymbol{\mu}}_{21}\right)^2+\left(f_j(\bar{X}_j)-\bar{\boldsymbol{\mu}}_{12}\right)^2}{\sigma^2}+B\right)\right| \\
\end{aligned}
\end{equation}
From Lagrange mean value theorem, we can know that, $|\exp (-x)-\exp (-y)|=\exp (-\xi)|y-x| \leq|x-y|$, where $\xi$ is between $x$ and $y$. Because of $
\left|\mathbf{P}\left(Y_i=c_1 \mid f_i(\bar{X}_i)\right)-\mathbf{P}\left(Y_j=c_1 \mid f_j(\bar{X}_j)\right)\right|>0 \text {, we can get } \frac{\left(f_j(\bar{X}_j)-\bar{\boldsymbol{\mu}}_{12}\right)^2+\left(f_j(\bar{X}_j)-\bar{\boldsymbol{\mu}}_{21}\right)^2}{\sigma^2}>B \text 
{ and set } x = \frac{\left(f_i(\bar{X}_i)-\bar{\boldsymbol{\mu}}_{11}\right)^2+\left(f_j(\bar{X}_j)-\bar{\boldsymbol{\mu}}_{22}\right)^2}{\sigma^2} - B$ and $y = \frac{\left(f_i(\bar{X}_i)-\bar{\boldsymbol{\mu}}_{21}\right)^2+\left(f_j(\bar{X}_j)-\bar{\boldsymbol{\mu}}_{12}\right)^2}{\sigma^2} - B$, which produces:
\begin{equation}
\begin{aligned}
& \left|\mathbf{P}\left(Y_i=c_1 \mid f_i(\bar{X}_i)\right)-\mathbf{P}\left(Y_j=c_1 \mid f_j(\bar{X}_j)\right)\right| \\
& \leq \frac{1}{\sigma^2}\left|\left[\left(f_i(\bar{X}_i)-\bar{\boldsymbol{\mu}}_{11}\right)^2-\left(f_j(\bar{X}_j)-\bar{\boldsymbol{\mu}}_{12}\right)^2\right]-\left[\left(f_i(\bar{X}_i)-\bar{\boldsymbol{\mu}}_{21}\right)^2-\left(f_j(\bar{X}_j)-\bar{\boldsymbol{\mu}}_{22}\right)^2\right]\right| \\
& = \frac{1}{\sigma^2} \left\lvert\,\left[\left(f_i(\bar{X}_i)-\frac{p^{(i)} \boldsymbol{\mu}_1+q^{(i)} \boldsymbol{\mu}_2}{p^{(i)}+q^{(i)}}\right)^2-\left(f_j(\bar{X}_j)-\frac{p^{(j)} \boldsymbol{\mu}_1+q^{(j)} \boldsymbol{\mu}_2}{p^{(j)}+q^{(j)}}\right)^2\right]\right. \\
& \left.-\left[\left(f_i(\bar{X}_i)-\frac{p^{(i)} \boldsymbol{\mu}_2+q^{(i)} \boldsymbol{\mu}_1}{p^{(i)}+q^{(i)}}\right)^2-\left(f_j(\bar{X}_j)-\frac{p^{(j)} \boldsymbol{\mu}_2+q^{(j)} \boldsymbol{\mu}_1}{p^{(j)}+q^{(j)}}\right)^2\right] \right\rvert\, \\
& = \left|\left(\frac{\left(p^{(i)}+p^{(j)}-q^{(i)}-q^{(j)}\right)\left(\boldsymbol{\mu}_2-\boldsymbol{\mu}_1\right)}{p^{(i)}+q^{(i)}} \right) \left(f_i\left(\bar{X}_i\right)-f_j\left(\bar{X}_j\right)+\frac{\left(p^{(i)}-p^{(j)}\right)\left(\boldsymbol{\mu}_2-\boldsymbol{\mu}_1\right)}{p^{(i)}+q^{(i)}}\right) \right|\\
& \leq \frac{\left\|\boldsymbol{\mu}_1-\boldsymbol{\mu}_2\right\|}{\sqrt{2 \pi} \sigma}\left(\left\|f_i(\bar{X}_i)-f_j(\bar{X}_j)\right\|+\left|\frac{p^{(i)}}{p^{(i)}+q^{(i)}}-\frac{p^{(j)}}{p^{(j)}+q^{(j)}}\right| \cdot\left\|\boldsymbol{\mu}_1-\boldsymbol{\mu}_2\right\|\right). \\
&
\end{aligned}
\end{equation}
which gives an upper bound. Note that $\left\|\boldsymbol{\mu}_1-\boldsymbol{\mu}_2\right\|$ and $\left|\frac{p^{(i)}}{p^{(i)}+q^{(i)}}-\frac{p^{(j)}}{p^{(j)}+q^{(j)}}\right|$ are the original feature and structure differences and we mark them as $a_1$ and $a_2$ respectively. Define $T_i \cap T_j$ = $T_a$, then we have:
\begin{equation}
\begin{aligned}
&\frac{\left\|\boldsymbol{\mu}_1-\boldsymbol{\mu}_2\right\|}{\sqrt{2 \pi} \sigma}\left(\left\|f_i(\bar{X}_i)-f_j(\bar{X}_j)\right\|+\left|\frac{p^{(i)}}{p^{(i)}+q^{(i)}}-\frac{p^{(j)}}{p^{(j)}+q^{(j)}}\right| \cdot\left\|\boldsymbol{\mu}_1-\boldsymbol{\mu}_2\right\|\right) \\
&= \frac{a_1}{\sqrt{2 \pi} \sigma}\left( \sum_{u \in T_a} \left[ \left(\bar{X}_{i u}-\bar{X}_{j u}\right)^2 \right] + \sum_{u \in T_i - T_j} \bar{X}_{i u}^2 + \sum_{u \in T_j - T_a} \bar{X}_{j u}^2\right) + \frac{a_1^2 a_2}{\sqrt{2 \pi} \sigma}
\end{aligned}
\end{equation}
To consider the effect of $\left|T_i \cap T_j\right|$ for the upper bound, we assume adding a new personalized feature index $u$ in it. We can compute changes $\Delta$ as follows:
\begin{equation}
\begin{aligned}
& \Delta = \frac{a_1}{\sqrt{2 \pi} \sigma}\left( \left(\bar{X}_{i u}-\bar{X}_{j u}\right)^2 - \bar{X}_{i u}^2 - \bar{X}_{j u}^2\right) \\
& = -\frac{2a_1 \bar{X}_{i u} \bar{X}_{j u}}{\sqrt{2 \pi} \sigma} \\
&< 0.
\end{aligned}
\end{equation}
Note that we get the personalized features through squeeze-and-excitation, so $\bar{X}_{i u}$ and $\bar{X}_{j u}$ have the same polarity as $\boldsymbol{\mu}_1$ and $\boldsymbol{\mu}_2$. Thus, with an increasing value of $\left|T_a\right|$, $\left|\mathbf{P}\left(Y_i=c_1 \mid f_i(\bar{X}_i)\right)-\mathbf{P}\left(Y_j=c_1 \mid f_j(\bar{X}_j)\right)\right|$ has a lower upper bound, and a larger $\bar{X}_{i u} \bar{X}_{j u}$ has a large effect, which finishes our proof.
\end{proof}

\clearpage
\subsection{Algorithm}
The detailed learning process of FPGC is illustrated in Algorithm \ref{Alg}.
\begin{algorithm}
\caption{FPGC}
\begin{algorithmic}[1]
\REQUIRE Input graph data $X$ and $\tilde{A}$; iteration number $\mathbf{K}$; hyperparameters $m$, $n$, $k$, $\lambda$.
\ENSURE The clustering result $\mathbf{R}$.
\STATE Calculate the normalized adjacency matrix $A = D^{-\frac{1}{2}}(\widetilde{A} + I)D^{-\frac{1}{2}}$.
\STATE Calculate filtered features $\bar{X}=A^kX$.
\STATE Randomly divide $\bar{X}$ into $m$ fields using Eq. (10).
\STATE Calculate $R$ using Eq. (11).
\FOR{$i = 1$ to $\mathbf{K}$}
    \STATE Obtain cluster-relavant features $X^n$ using Eq. (1)-(3).
    \STATE Obtain the first output using Eq. (9).
    \STATE Perform data augmentation based on feature cross through Eq. (12).
    \STATE Obtain the second output using Eq. (13).
    \STATE Calculate the contrastive learning loss $\mathcal{L}_{con}$ Eq. (14).
    \STATE Calculate the reconstruction loss $\mathcal{L}_{re}$ with Eq. (15).
    \STATE Update node weights across each output by minimizing $\mathcal{L}$ in Eq. (16).
\ENDFOR
\STATE Perform K-means on $\frac{Y+Y^{\prime}}{2}$ to obtain clustering result  $\mathbf{R}$.
\RETURN $\mathbf{R}$.
\end{algorithmic}
\label{Alg}
\end{algorithm}